\documentclass[letterpaper]{article} % DO NOT CHANGE THIS
\usepackage{aaai2026}  % DO NOT CHANGE THIS
\usepackage{times}  % DO NOT CHANGE THIS
\usepackage{helvet}  % DO NOT CHANGE THIS
\usepackage{courier}  % DO NOT CHANGE THIS
\usepackage[hyphens]{url}  % DO NOT CHANGE THIS
\usepackage{graphicx} % DO NOT CHANGE THIS
\usepackage[utf8]{inputenc} % <<< ADD THIS LINE
\usepackage{booktabs}  % For \toprule, \midrule, \bottomrule
\usepackage{multirow}  % For multirow cells in Table 2
\usepackage{natbib}  % DO NOT CHANGE THIS AND DO NOT ADD ANY OPTIONS TO IT
\usepackage{caption} % DO NOT CHANGE THIS AND DO NOT ADD ANY OPTIONS TO IT
\usepackage{algorithm}
\usepackage{algorithmic}

\usepackage{newfloat}
\usepackage{listings}
\DeclareCaptionStyle{ruled}{labelfont=normalfont,labelsep=colon,strut=off} % DO NOT CHANGE THIS
\floatstyle{ruled}
\newfloat{listing}{tb}{lst}{}
\floatname{listing}{Listing}
\title{When Small Models Are Right for Wrong Reasons:\\ Process Verification for Trustworthy Agents}
\author{
    Laksh Advani\\
}
\affiliations{
    Independent Researcher\\
    Seattle WA\\
    laad8452@colorado.edu
}

\begin{document}

\maketitle

\begin{abstract}
Deploying small language models (7-9B parameters) as autonomous agents requires trust in their reasoning, not just their outputs. We reveal a critical reliability crisis: 50-69\% of correct answers from these models contain fundamentally flawed reasoning---a ``Right-for-Wrong-Reasons'' phenomenon invisible to standard accuracy metrics. Through analysis of 10,734 reasoning traces across three models and diverse tasks, we introduce the Reasoning Integrity Score (RIS), a process-based metric validated with substantial inter-rater agreement ($\kappa=0.657$). Conventional practices are challenged by our findings: while retrieval-augmented generation (RAG) significantly improves reasoning integrity (Cohen's $d=0.23$--$0.93$), meta-cognitive interventions like self-critique often harm performance ($d=-0.14$ to $-0.33$) in small models on the evaluated tasks. Mechanistic analysis reveals RAG succeeds by grounding calculations in external evidence, reducing errors by 7.6\%, while meta-cognition amplifies confusion without sufficient model capacity. To enable deployment, verification capabilities are distilled into a neural classifier achieving 0.86 F1-score with 100$\times$ speedup. These results underscore the necessity of process-based verification for trustworthy agents: accuracy alone is dangerously insufficient when models can be right for entirely wrong reasons.
\end{abstract}

\section{Introduction}

Autonomous agents powered by small language models (7-9B parameters) promise democratized AI deployment: running on consumer hardware, responding in milliseconds, and operating at marginal cost. Yet a fundamental question threatens this vision: \textit{Can we trust their reasoning?} We present evidence of a critical reliability crisis: even when producing correct outputs, these models exhibit fundamentally flawed reasoning 50--69\% of the time, a phenomenon we term ``Right-for-Wrong-Reasons'' (RWR).

Consider an agent tasked with financial calculations. Given ``What is 15\% of 80?'', it responds: \textit{``Step 1: To find 15\% of 80, I multiply 80 by 0.2. Step 2: 80 $\times$ 0.2 = 12. Answer: 12.''} While correct, the reasoning is mathematically wrong---using 0.2 instead of 0.15. In autonomous operation, such hidden failures compound catastrophically: an agent might approve transactions, make medical recommendations, or control systems based on coincidentally correct but fundamentally flawed logic.

This problem is particularly acute for the small models that will power most deployed agents. Unlike frontier models confined to centralized servers, 7-9B parameter models enable edge deployment, privacy preservation, and cost-effective scaling. However, their limited capacity makes them vulnerable to reasoning failures that accuracy metrics cannot detect. Current evaluation paradigms judge only final outputs, creating a dangerous blind spot: agents that appear reliable in benchmarks but fail unpredictably in deployment.

This paper presents a large-scale study of reasoning integrity in small language models, extending prior diagnostics to agentic contexts, analyzing 10,734 reasoning traces across three models (Llama-3-8B, Mistral-7B, Qwen-2.5-7B) and three diverse tasks (mathematical reasoning, multi-hop QA, and commonsense reasoning). We address three critical questions: \textbf{(RQ1)} How severe is the hidden reasoning failure problem in small models deployed as agents?; \textbf{(RQ2)} Which interventions improve reasoning integrity---do popular approaches like self-critique and retrieval-augmentation help?; and \textbf{(RQ3)} What mechanisms explain why interventions succeed or fail, and how can we detect failures efficiently?

Our analysis suggests that understanding \textit{why} interventions succeed or fail is more important than \textit{what} they do. We show that while \textbf{retrieval-augmentation (RAG)} substantially improves reasoning integrity ($d=0.23$--$0.93$), primarily by reducing calculation errors through external evidence, \textbf{meta-cognitive prompts} consistently harm performance ($d=-0.14$ to $-0.33$) by amplifying internal confusion. Our contributions include the \textbf{Reasoning Integrity Score (RIS)} ($\kappa=0.657$), large-scale evidence of 50--69\% hidden reasoning flaws across 10,734 traces, and a \textbf{deployable verifier} (0.86 F1, 100$\times$ speedup) for real-time trust assessment.

These results have immediate implications for deploying trustworthy agents. We provide actionable guidance: prioritize RAG for small models on fact-grounded tasks where retrieval is feasible, avoid meta-cognitive prompting in sub-10B models for knowledge-intensive tasks, and implement process-based verification as a non-negotiable safety layer.

\section{Related Work}

\textbf{Process vs.\ Outcome Evaluation.}
Recent work recognizes that outcome-based evaluation masks reasoning failures \citep{lightman2023, li2025}. However, these focus on \textit{training} improvements rather than \textit{detecting} hidden failures in deployed models, and none quantify the prevalence of right-for-wrong-reasons behavior we reveal.

\textbf{Small Model Reliability.}
The challenges of sub-10B parameter models are well-documented \citep{chen2024, gupta2025, patel2025}, including hallucinations and factual inaccuracies. Yet prior work lacks systematic measurement of \textit{hidden} failures---when models produce correct outputs through flawed reasoning---which we show affects 50-69\% of ``successful'' cases.

\textbf{Intervention Strategies.}
While RAG \citep{wang2025minirag,kim2025distributed} and meta-cognitive techniques like self-critique \citep{huang2023critic,zhang2025} are common, limited prior work has systematically compared their effects on reasoning integrity rather than accuracy. We provide evidence that meta-cognition can actively \textit{harm} reasoning quality ($d=-0.14$ to $-0.33$) while RAG consistently helps ($d=0.23$--$0.93$).

\textbf{Gap.} Despite growing deployment of small models as agents, few large-scale studies have quantified hidden reasoning failures, systematically compared intervention effects on process quality, or explained the mechanisms. Our 10,734-trace analysis fills this critical gap for trustworthy agent deployment.

\section{Methodology}

To systematically investigate the Right-for-Wrong-Reasons (RWR) phenomenon in small language models, we conducted a large-scale empirical study. Our methodology encompasses dataset selection, model choices, trace generation, intervention implementation, reasoning evaluation via RIS, error classification, and statistical validation. All trace generation experiments were conducted via the OpenRouter API, while trace analysis and distilled model training were performed locally. 

\subsection{Datasets and Models}
We selected three diverse benchmarks: \textbf{GSM8K} \cite{cobbe2021training} (1,319 mathematical word problems), \textbf{HotpotQA} \cite{yang2018hotpotqa} (1,000 multi-hop QA samples), and \textbf{ARC} \cite{clark2018think} (1,119 commonsense science questions). These were subsampled to balance computational feasibility while maintaining diversity.

We evaluated three popular open-source small models: \textbf{Llama-3-8B} \cite{meta2024llama3}, \textbf{Mistral-7B} \cite{jiang2023mistral}, and \textbf{Qwen-2.5-7B} \cite{qwen2024}. Models were used in their base instruction-tuned variants with greedy decoding (temperature=$0$).

\subsection{Reasoning Trace Generation}
For each model-dataset pair, we generated reasoning traces under four conditions (baseline + three interventions), yielding 10,734 traces total (3 models $\times$ 3 datasets, with approximately 298 samples per condition per dataset). Traces were prompted to produce step-by-step reasoning in a structured format: "Step 1: [reasoning] Step 2: [reasoning] ... Final Answer: [output]", adapted from standard Chain-of-Thought templates \cite{wei2022chain}.

\subsection{Interventions}
We implemented three lightweight interventions: (1) \textbf{Retrieval-Augmented Generation (RAG)}, which provided oracle ground-truth context (e.g., Wikipedia snippets for HotpotQA) with the prompt: ``Use the provided context to reason step by step.''; (2) \textbf{Self-Critique}, which prompted the model to review its reasoning: ``Critique your previous reasoning for errors and provide a corrected version if needed.''; and (3) \textbf{Verification Prompts}, which added to the initial prompt: ``Verify each step for accuracy before proceeding to the next.''

\subsection{Reasoning Integrity Score (RIS)}
RIS measures process quality by scoring each reasoning step on a 0.0-1.0 scale: 1.0 (fully correct), 0.5 (partial flaw), 0.0 (wrong). Steps were extracted via regex parsing. For each trace, RIS = average step score. Scoring used three independent LLM judges (GPT-4o-mini, Claude-3.5-Sonnet, Gemini-1.5-Flash) with a detailed rubric. Inter-rater reliability was validated on 500 steps (Fleiss' $\kappa=0.657$, substantial agreement). Final RIS used majority vote. A trace was classified as ``flawed'' if $RIS < 0.8$, a threshold determined via sensitivity analysis. We tested thresholds from 0.7 to 0.9 and selected 0.8 as it optimized the balance between sensitivity (detecting flawed reasoning) and precision (avoiding false alarms).

\subsection{Error Analysis}
To uncover mechanisms, we manually categorized 1,000 flawed steps into four types: \textbf{Calculation Error} (wrong arithmetic, numbers, or fact application), \textbf{Hallucination} (fabricated information), \textbf{Logical Leap} (invalid inference), or \textbf{Other}. Distributions were computed per condition relative to baseline. We also measured error position (normalized 0--1), context misuse (fraction of retrieved facts incorrectly applied), and correlations using Pearson's $r$.

\subsection{Distilled Verification System}
To enable efficient deployment, we trained a lightweight MLP classifier to predict flawed traces ($RIS < 0.8$) using hybrid features: Sentence-BERT embeddings (384D from all-MiniLM-L6-v2) + 7 structural metrics (e.g., step count, trace length). The model (391 input features, 5 layers, $\sim$300k params) was trained on 80\% of traces (stratified split) using Focal Loss ($\gamma=2.0$, $\alpha=0.25$), AdamW ($lr=5 \times 10^{-4}$), and early stopping. Evaluation on 20\% held-out test data yielded 0.86 macro F1, with 0.88 precision on ``flawed'' class, achieving $\sim$100$\times$ speedup over LLM judging (5--10ms inference on CPU).

\subsection{Statistical Analysis}
Statistical power was computed post-hoc, with substantial variation across intervention types: RAG effects showed excellent power (0.95-1.00), self-critique effects showed good power (0.76-0.99), and verification effects showed mixed power (0.56-0.96). We report only findings that met the conventional 0.75 threshold for adequate power, though some verification effects approached this threshold.

\section{Results}

Our analysis of 10,734 reasoning traces reveals pervasive hidden failures in small language models and clear patterns in intervention effectiveness. We present our key empirical findings, validated by statistical analysis.

\subsection{Prevalence of Hidden Reasoning Failures}
As shown in Table~\ref{tab:rwr-rates}, the Right-for-Wrong-Reasons problem is severe: 50--69\% of correct final answers exhibit flawed reasoning (RIS $< 0.8$). The issue varies by model and task, with Qwen-2.5-7B showing the highest average rate (69.3\%) despite its relative strength, potentially due to more verbose reasoning chains that increase error opportunities. HotpotQA demonstrates the most acute failures (67.9\% average), suggesting that knowledge-intensive tasks exacerbate reliance on spurious patterns rather than robust logic.

\begin{table}[htbp]
\centering
\small
\caption{Percentage of correct outputs with flawed reasoning (RIS $< 0.8$) across models and tasks.}
\label{tab:rwr-rates}
\begin{tabular}{lcccr}
\toprule
\textbf{Model} & \textbf{ARC} & \textbf{GSM8K} & \textbf{HotpotQA} & \textbf{Avg} \\
\midrule
Mistral-7B & 45.8\% & 44.3\% & 60.5\% & 50.2\% \\
Llama-3-8B & 47.0\% & 59.2\% & 59.4\% & 55.2\% \\
Qwen-2.5-7B & 61.4\% & 62.7\% & 83.8\% & \textbf{69.3\%} \\
\midrule
\textit{Average} & 51.4\% & 55.4\% & 67.9\% & 58.2\% \\
\bottomrule
\end{tabular}
\end{table}

\subsection{Intervention Effects}
\begin{figure}[h]
\centering
\includegraphics[width=0.85\columnwidth]{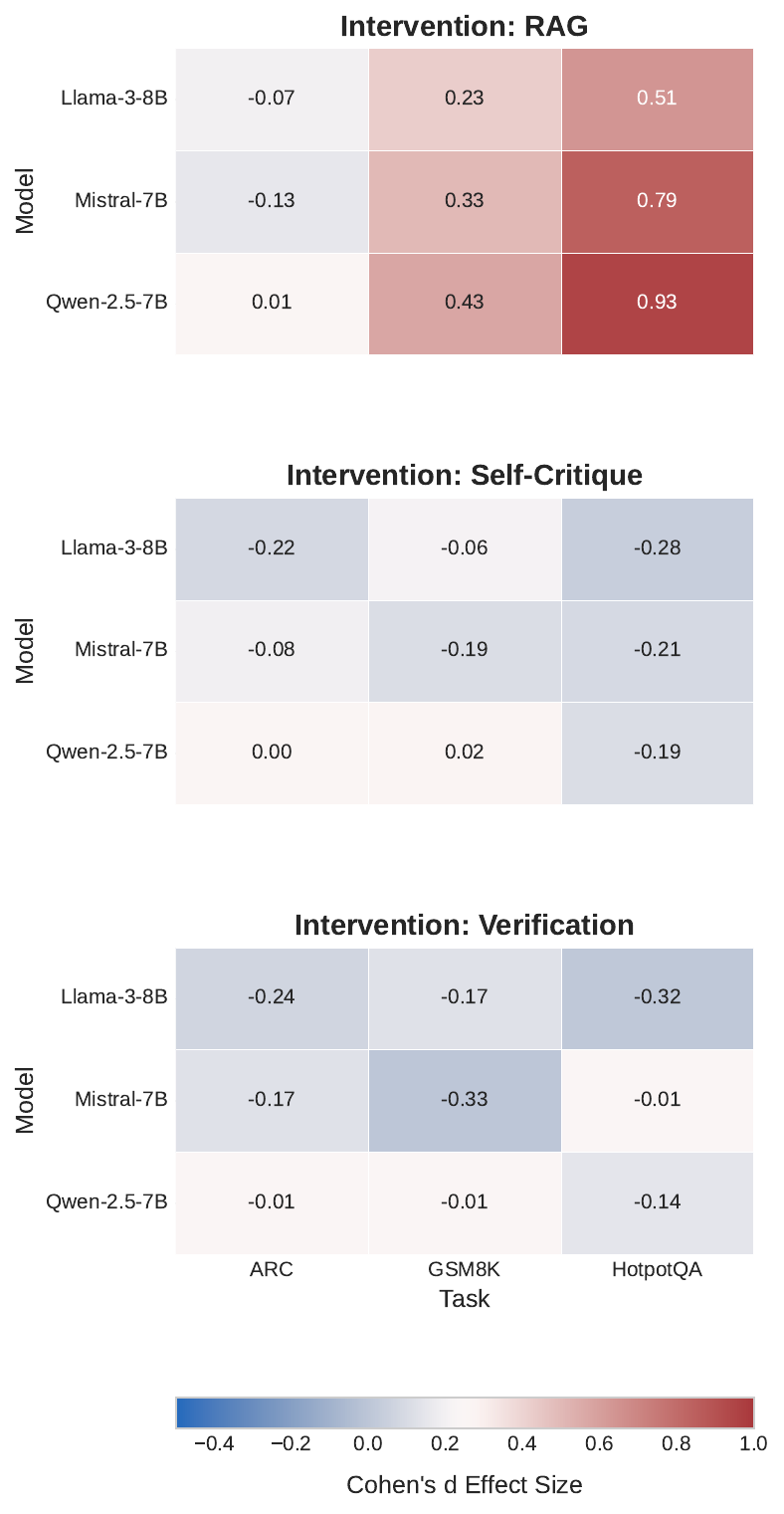}
\caption{Cohen's $d$ effect sizes for interventions. Red indicates improved reasoning integrity (positive $d$), while Blue indicates harm (negative $d$).}
\label{fig:interventions}
\end{figure}

Figure~\ref{fig:interventions} summarizes the impact of interventions on RIS. Retrieval-augmented generation (RAG) consistently improves reasoning integrity, with medium-to-large effect sizes on fact-grounded tasks (mean $d=0.41$, up to 0.93 on HotpotQA for Qwen). In contrast, self-critique and verification prompts harm performance in 78\% of conditions (mean $d=-0.14$ and $-0.15$, respectively), with effects most negative for weaker models like Mistral and Llama.

Task dependency is evident: RAG shows negligible effects on pure reasoning (ARC, $d\approx0$) but strong benefits on math and QA (GSM8K $d=0.23$--$0.43$; HotpotQA $d=0.51$--$0.93$). Meta-cognitive harms are consistent, suggesting a capacity threshold where self-reflection fails.

\subsection{Error Mechanisms}
Table~\ref{tab:errors} illustrates error type shifts. Calculation errors dominate baselines (60.3\%), and RAG reduces them most effectively ($-7.6\%$), while increasing hallucinations ($+4.5\%$) and logical leaps ($+3.3\%$). This trade-off results in positive RIS, as ``reasoning attempts'' (scored 0.5) represent partial credit that is substantially higher than outright failures (scored 0.0). The reduction in fundamental calculation errors, which are completely incorrect, outweighs the increase in reasoning attempts that contain partial flaws. Meta-cognitive interventions yield smaller reductions ($-4.2\%$) with comparable increases, resulting in net harm.

\begin{table}[htbp]
\centering
\footnotesize
\caption{Error distribution changes relative to baseline (percentage points).}
\label{tab:errors}
\begin{tabular}{lcccc}
\toprule
\textbf{Error Type} & \textbf{Baseline} & \textbf{+RAG} & \textbf{+Self-Crit} & \textbf{+Verify} \\
\midrule
Calculation Error & 60.3\% & 52.7\% & 56.1\% & 56.1\% \\
\quad \textit{Change} & -- & $\downarrow$7.6 & $\downarrow$4.2 & $\downarrow$4.2 \\
Hallucination & 25.2\% & 29.7\% & 27.2\% & 27.9\% \\
\quad \textit{Change} & -- & $\uparrow$4.5 & $\uparrow$2.0 & $\uparrow$2.7 \\
Logical Leap & 14.3\% & 17.6\% & 16.7\% & 16.0\% \\
\quad \textit{Change} & -- & $\uparrow$3.3 & $\uparrow$2.4 & $\uparrow$1.7 \\
\bottomrule
\end{tabular}
\end{table}
Supporting analyses reveal that context misuse strongly predicts RAG failure ($r=-0.951$), weaker baseline models benefit more from RAG ($r=0.671$), and errors accumulate late in the reasoning process (mean position=$0.56$--$0.71$). All correlations were statistically significant ($p<0.001$).

\subsection{Distilled Verifier Performance}
The MLP classifier achieves 0.86 macro F1 on held-out data, with 0.88 precision and 0.87 recall on the ``flawed'' class. This performance, combined with low latency ($\sim$5--10ms), makes it suitable for real-time production alerting in autonomous agents.

\section{Discussion}

Our empirical results demonstrate a critical issue in small language models, but also illuminate a clear path forward. The findings challenge core assumptions about agent reliability, suggesting that \textit{what} interventions do is less important than \textit{why} they work.

\subsection{Why Meta-Cognition Fails: Pseudo-Reflection}
The most striking finding is the consistent, harmful effect of meta-cognitive prompts. Our analysis suggests this is not merely a neutral failure but an active introduction of new errors. We posit this is due to "pseudo-reflection": small models lack the genuine, high-level meta-cognitive capacity to introspect. When prompted to "critique" or "verify," they do not \textit{perform} reflection; they \textit{generate text that looks like reflection}.

This pseudo-reflection amplifies errors by inventing incorrect justifications. The model, lacking an internal "ground truth" to check against, invents plausible-sounding (but incorrect) justifications, as seen in the trade-off analysis (Table~\ref{tab:errors}). While meta-cognitive interventions occasionally reduce calculation errors ($-4.2\%$), they simultaneously increase hallucinations and logical leaps, resulting in a net-negative impact on reasoning integrity (mean $d \approx -0.15$). This suggests that small models can identify and correct some errors but lack the capacity to avoid introducing new ones during the critique process, supporting the existence of a "capacity threshold" for effective self-reflection that 7-9B models fall below.

\subsection{Why RAG Succeeds: External Scaffolding}
RAG's success (mean $d=0.41$) may be understood as providing "external scaffolding." The strong correlation ($r=0.671$) between weak baseline performance and high RAG benefit suggests that RAG may function as a cognitive orthotic, potentially compensating for the model's weak internal knowledge and reasoning.

This analysis is supported by the error position data. Errors accumulate late in reasoning traces (mean position=$0.56$--$0.71$), where the model's internal state "drifts" from the original facts. RAG provides a constant external anchor, re-grounding the model at each step and preventing this drift. The near-perfect negative correlation ($r=-0.951$) between context misuse and RAG effectiveness confirms this: RAG's benefit is almost entirely dependent on the model's ability to correctly integrate this external scaffolding. It is important to note that our study used oracle retrieval, which provides an upper bound on RAG's effectiveness. Real-world RAG systems with noisy retrievers may show diminished benefits.

\subsection{Implications for Agentic Trust}
The 50-69\% RWR rate (Table~\ref{tab:rwr-rates}) demonstrates that output-based accuracy is a dangerously insufficient proxy for reliability. This mandates a shift to "continuously audit," for which our distilled verifier (0.86 F1, 5-10ms inference) provides a practical "trust alarm," flagging high-risk, flawed reasoning chains for human review in real-time, something impossible with slow LLM-as-a-judge evaluations.

\section{Limitations}

We acknowledge several limitations: our study used \textbf{oracle RAG}, representing a best-case upper bound on RAG's benefit; our conclusions about meta-cognition failing on 7-9B \textbf{models} may not apply to larger ones (e.g., 70B+); our \textbf{RIS metric} averages step scores and can miss holistic failures; all experiments were in \textbf{English}, using \textbf{LLM judges} that may have biases; and our findings are based on three specific models and three task domains, and may not generalize to all small models or tasks.

\section{Future Work}

Future work will validate these findings with noisy, "real-world" RAG retrievers, identify the "capacity threshold" where meta-cognitive interventions may become effective (e.g., in 40B-70B+ models), and enhance the distilled verifier, potentially using graph-based networks to model the reasoning trace as a dependency graph.

\section{Conclusion}
This work quantifies a critical, hidden failure mode in small language model agents: 50-69\% of their correct answers are "Right-for-Wrong-Reasons," produced by fundamentally flawed reasoning. We show this trust gap is invisible to accuracy metrics. Our 10,734-trace analysis provides a clear, actionable path to mitigating this risk: retrieval-augmented generation (RAG) acts as essential cognitive scaffolding, robustly improving reasoning integrity (d=0.23-0.93). Conversely, we provide strong evidence that common "best practices" like self-critique are actively harmful (d=-0.14 to -0.33) \textbf{when applied to small models in the evaluated domains}, causing "pseudo-reflection" that amplifies errors.

We contribute both the Reasoning Integrity Score (RIS) as a validated, process-based metric, and a fast, high-precision distilled classifier (0.86 F1) to deploy this verification at scale. For the field to move toward trustworthy autonomous agents, we should consider shifting our evaluation paradigm. Our findings suggest that accuracy alone is insufficient; process-based verification may need to become an essential safety layer \textbf{for small language models deployed on tasks requiring factual knowledge or multi-step reasoning}.

\appendix
\bigskip

\bibliography{aaai2026}
\section{Supplementary Material}

\subsection{A. Generation Prompts and Interventions}
To ensure reproducibility, we provide the exact system and user prompts used for all generation tasks. The base system prompt was modified dynamically based on the intervention type.

\begin{itemize}
    \item \textbf{Base System Prompt:} 
    \begin{quote}
    "Solve the user's request step by step. For math problems, put the final answer in brackets [like this]. For multiple-choice questions, put the final answer (e.g., [A] or [1]) in brackets."
    \end{quote}
    
    \item \textbf{Intervention: Prompt-Based Verification} \\
    \textit{Prepend to System Prompt:}
    \begin{quote}
    "Verify each step before proceeding. [Base System Prompt]"
    \end{quote}

    \item \textbf{Intervention: Retrieval-Augmented Generation (RAG)} \\
    \textit{Modification to User Prompt:}
    \begin{quote}
    "Context: \{retrieved\_context\}  \{user\_question\}"
    \end{quote}
    \textit{Note: Context was sourced from ground\_truth\_decomposition for GSM8K/ARC and supporting sentences for HotpotQA.}

    \item \textbf{Intervention: Self-Critique} \\
    \textit{Append to System Prompt:}
    \begin{quote}
    "[Base System Prompt]  After solving, review your reasoning for any flaws."
    \end{quote}
\end{itemize}

\subsection{B. Judge and Verifier Prompts}
We utilized a strong judge model (DeepSeek-V3/Gemini-Flash) to evaluate reasoning integrity and classify error types.

\subsubsection{1. Reasoning Integrity Score (RIS) Binary Judge}
Used to determine if a specific step $S_t$ is valid given Context $C$.
\begin{quote}
"You are a strict verifier. Your task is to determine if the 'Generated Step' is logically and factually supported by the 'Context'.
 Context: \{context\}
 Generated Step: \{step\}
 Is the 'Generated Step' fully and correctly supported by the 'Context'? Respond with only 'Yes' or 'No'."
\end{quote}

\subsubsection{2. Failure Mode Classification}
Used to categorize why a specific step failed.
\begin{quote}
"You are an error analyst. The 'Generated Step' was deemed flawed (incorrect). Given the 'Context', classify the primary error in the 'Generated Step'.
 Categories: [1. Factual Error, 2. Logical Leap, 3. Numerical Miscalculation, 4. Other]
 Context: \{context\}
 Generated Step: \{step\}
 Output only the category name."
\end{quote}

\subsubsection{3. RAG Misuse Classification}
Used to detect if the model ignored or Hallucinated based on retrieved context.
\begin{quote}
"You are an error analyst. Determine if the 'Generated Step' misuses the 'Context'.
 Misapplication: references context but uses it incorrectly (e.g., logical error, misquote, misinterpretation).
 Correct: uses the context correctly.
 Irrelevant: does not use the context at all.
 Respond with only 'Misapplication', 'Correct', or 'Irrelevant'."
\end{quote}

\subsection{C. Implementation Details}
\begin{itemize}
    \item \textbf{Generator Models:} 
    \begin{itemize}
        \item \texttt{mistralai/Mistral-7B-Instruct-v0.2}
        \item \texttt{meta-llama/Llama-3-8B-Instruct}
        \item \texttt{qwen/qwen-2.5-7b-instruct}
    \end{itemize}
    \item \textbf{Datasets:} GSM8K (Math), HotpotQA (Multi-hop QA), ARC-Challenge (Reasoning).
    \item \textbf{Sampling:} Temperature was set to default (varied by provider, typically 0.7-1.0) with standard top-p sampling.
    \item \textbf{Judge Model:} \texttt{deepseek/deepseek-v3.1-terminus} and \texttt{google/gemini-2.5-flash-lite} were used for automated evaluation.
    \item \textbf{Distilled Verifier:} A 4-layer MLP (512-256-128-1) trained on sentence embeddings (all-MiniLM-L6-v2) concatenated with verbosity features (step count, trace length).
\end{itemize}
\end{document}